# A Generative Process for Sampling Contractive Auto-Encoders

Salah Rifai[(1)]                                                             RIFAISAL@IRO.UMONTREAL.CA
**Yoshua Bengio**[(1)]                                      BENGIOY@IRO.UMONTREAL.CA
**Yann N. Dauphin**[(1)]                                DAUPHIYA@IRO.MONTREAL.CA
**Pascal Vincent**[(1)]                                     VINCENTP@IRO.UMONTREAL.CA

[(1)] Dept. IRO, Université de Montréal. Montréal (QC), H3C 3J7, Canada

## Abstract

The contractive auto-encoder learns a representation of the input data that captures the local manifold structure around each data point, through the leading singular vectors of the Jacobian of the transformation from input to representation. The corresponding singular values specify how much local variation is plausible in directions associated with the corresponding singular vectors, while remaining in a high-density region of the input space. This paper proposes a procedure for generating samples that are consistent with the local structure captured by a contractive auto-encoder. The associated stochastic process defines a distribution from which one can sample, and which experimentally appears to converge quickly and mix well between modes, compared to Restricted Boltzmann Machines and Deep Belief Networks. The intuitions behind this procedure can also be used to train the second layer of contraction that pools lower-level features and learns to be invariant to the local directions of variation discovered in the first layer. We show that this can help learn and represent invariances present in the data and improve classification error.

## 1. Introduction

A central objective of machine learning is to generalize from training examples to new configurations of the observed variables, and in the most general setup this means deciding how to redistribute the probability mass associated with each training example from the empirical distribution. Classical non-parametric density estimation does this by convolving the empirical distribution with a smoothing kernel such as the Gaussian (giving rise to the Parzen density estimator). This spreads each point mass into the nearby volume surrounding each training point, but unfortunately it does so isotropically (in the same way in all directions around each training point). This means that if the true density tends to concentrate near low-dimensional manifolds (this is called the *manifold hypothesis* (Cayton, 2005; Narayanan and Mitter, 2010; Rifai et al., 2011a)), then a lot of densely packed training examples will be required to achieve a high model density near the manifold and a low model density away from it.

Whereas Principal Components Analysis (PCA) discovers a linear manifold near which the density may concentrate, *Non-Linear Manifold Learning* algorithms (Schölkopf et al., 1998; Roweis and Saul, 2000; Tenenbaum et al., 2000) attempt to discover the *structure* of manifolds (which may be non-linear) near which the true density concentrates. In some cases a manifold learning algorithm can be generalized to obtain a model that tells us (explicitly or implicitly) how to allocate probability mass everywhere, and not just identify the manifold. For example, PCA corresponds to a Gaussian model, where the variances in directions orthogonal to the manifold are small and constant (across all these directions), and this can be extended to non-linear structures with a mixture model (Tipping and Bishop, 1999) or the kernel trick (Schölkopf et al., 1998).

In many manifold learning algorithms, the local shape of the manifold is specified by a local basis which indicates the plausible directions of variation, i.e., the tangent plane at any point on the manifold. Different algorithms propose different ways of stitching these lo-cal planes or local pancakes in order to construct a global manifold structure or a global density. However, a potentially serious limitation of most manifold learning algorithms is that they are based on *local generalization*: they infer these local tangent planes based mostly on the training examples in the *neighborhood* of the point of interest. As discussed in Bengio and Monperrus (2005) and Bengio et al. (2006b), this raises a curse of dimensionality issue: if the manifold of interest has many ups and downs then the number of examples required to capture the manifold increases exponentially with manifold dimension and the manifold curvature. The model cannot generalize in an up or down of the manifold for which there are no examples to map out its variations. A first non-linear manifold learning algorithm with non-local generalization was proposed by Bengio et al. (2006a), stimulating the work presented here, which also follows up on more recent work in the area of Deep Learning (Hinton et al., 2006; Bengio, 2009), more specifically the Contractive Auto-Encoder (CAE) (Rifai et al., 2011b;a), described in more detail in the next section. The CAE is a representation-learning algorithm which also captures local manifold structure and has the potential for non-local generalization.

The first main contribution of this paper (section 3) is a novel way to use the CAE to construct a generative procedure which implicitly specifies a density concentrating near the captured manifolds. The second main contribution of this paper (section 4) is inspired by the ideas of the first and consists in a novel way to *train* a pooling layer on top of the features learned by a CAE, so as to make them *invariant* to the local directions of variation captured by the lower-level CAE features. Another contribution of this paper regards the training of a multi-layer CAE, in particular using the above trained pooling layer, whereas previous CAE papers involved only single-layer CAEs, possibly stacked to get deeper representations. Section 5 shows results from the proposed generative process and from the invariance-seeking pooling algorithm for the CAE, illustrating the advantages brought by these two contributions.

## 2. Characterizing a data manifold with a Contractive Auto-Encoder

Deep Learning algorithms learn a representation of input data $x$, which is typically used either to construct a classifier[1] or in order to capture the structure of $P(x)$. They are *deep* if these representations have multiple levels, and the number of levels is a hyperparameter that can be chosen in a data-dependent way. Typically, higher levels are defined in terms of lower levels and are expected to represent more abstract features, e.g., better capturing structure present in the input distribution such as invariance (Goodfellow et al., 2009) or manifold structure. See Bengio (2009) for a review of Deep Learning algorithms. A major breakthrough in this area occurred in 2006 (Hinton et al., 2006) with the successful idea that deep architectures could be trained by stacking single-level unsupervised representation learning algorithms such as the Restricted Boltzmann Machine (RBM) or auto-encoder and sparse coding variants.

The Contractive Auto-Encoder (CAE) is an unsupervised feature learning algorithm that has been successfully applied in the training of deep networks, i.e., CAE layers can be stacked to form deeper representations. Its parametrization is essentially the same as that of a Restricted Boltzmann Machine, but contrary to RBMs its training procedure is deterministic, and consists in minimizing, through gradient descent, a simple objective that can be efficiently computed analytically and exactly.

In this section, we briefly review the CAE algorithm, its interpretation as modeling a data manifold, and how one can use a trained CAE to extract the local tangent space to that manifold at a point.

### 2.1. The Contractive Auto-Encoder training objective

Let us briefly introduce the notation that we will be using throughout, and formalize the operations of the CAE algorithm, following Rifai et al. (2011b).

From an input $x \in [0,1]^d$, a $k$-dimensional feature vector is computed as a hidden layer, e.g.,

$$h = f(x) = s(Wx + b_h), \qquad (1)$$

where $W \in \mathbb{R}^{k \times d}$ and $b_h \in \mathbb{R}^k$ are parameters of the model, and $s$ is the element-wise logistic sigmoid $s(z) = \frac{1}{1+e^{-z}}$. From hidden representation $h$, a *reconstruction* of $x$ is obtained as

$$r = g(f(x)) = s(W^T f(x) + b_r),$$

where $b_r \in \mathbb{R}^d$ is the reconstruction bias vector. Although the encoder $f(\cdot)$ and decoder $g(\cdot)$ can themselves have multiple layers, previous work on the CAE has focused on training only single-layer CAEs (and then stacking them to form deeper representations).

A *reconstruction loss* $L(x, r)$ measures how well the input is reconstructed from the hidden representation.

---

[1] or other supervised predictors such as $P(y|x)$ for some target variable $y$

Following Rifai et al. (2011b), we will be using a cross-entropy loss:

$$L(x, r) = -\sum_{i=1}^{d} x_i \log(r_i) + (1 - x_i) \log(1 - r_i).$$

The set of parameters of this model is $\theta = \{W, b_h, b_r\}$. The training objective being minimized in a traditional auto-encoder is simply the average reconstruction error over a training set $\mathcal{D}$. The Contractive Auto-Encoder adds a regularization term to this objective, that penalizes the *sensitivity* of the features to the input, measured as the Frobenius norm of the Jacobian matrix $J(x) = \frac{\partial f(x)}{\partial x}$.

In summary, the parameters $\theta$ of the CAE are learned by minimizing:

$$\mathcal{J}_{\text{CAE}}(\theta; \mathcal{D}) = \sum_{x \in \mathcal{D}} \left( L(x, g(f(x))) + \lambda \|J(x)\|^2 \right) \quad (2)$$

where $\lambda$ is a non-negative regularization hyper-parameter that controls how strongly the norm of the Jacobian is penalized.

### 2.2. How the CAE models a data manifold

A useful non-parametric modeling hypothesis for high dimensional data with a complicated structure, such as natural images or sounds, is the so-called *manifold hypothesis* (Cayton, 2005; Narayanan and Mitter, 2010; Rifai et al., 2011a). It hypothesizes that, while in its raw representation such data may appear to live in a high dimensional space, in reality its probability density is likely to be relatively high only along stripes of a much lower-dimensional non-linear sub-manifold, embedded in this high-dimensional Euclidean space[2]. This should not be taken to mean that all data must lie strictly on said sub-manifold (that we will from now on simply call manifold), but merely that the probability density is expected to decrease very rapidly when moving away from it. Since this manifold is thought to support the high data density regions, capturing and exploiting its precise structure and location within the high dimensional space is viewed as a key to better generalization.

To interpret the CAE from this perspective, as is done in Rifai et al. (2011b), it is useful to think of an autoencoder as projecting an input point $x$ onto a low dimensional manifold, and of the hidden representation as the coordinates of that projection in a coordinate system within the manifold. One may view this as a non-linear generalization of PCA, where PCA would correspond to projecting onto a linear sub-manifold. Let us consider a point $x$ and the manifold-coordinates of its projection (i.e. its hidden representation $h(x)$), and how they change as we slightly move $x$. For movements "parallel" to the manifold (i.e. in directions within the *tangent space* at the projection), the projection and thus $h$ will follow suit. Ideally, small movements "orthogonal" to the manifold however should not change the projection or $h$.

Now CAEs' contractive penalty pressures the hidden representation not to change when moving $x$, whatever the direction (it is an isotropic pressure). But this is counterbalanced in the CAE training criterion by the need to correctly reconstruct training points, so that $h$ *has to* be sensitive at least to moves in directions that lead to other likely points (such as training set neighbors). As a result, of this trade-off, the CAE's hidden representation will learn, at training points, to be sensitive only to movements alongside the manifold of high density (i.e. spanning the tangent space to the manifold), because these can yield points it must be able to reconstruct well, while sensitivity to all other directions (those orthogonal to the manifold) will have been shrunk.

This interpretation is supported in Rifai et al. (2011b) by an empirical analysis of the sensitivity of $f(x)$ to input directions: a Singular Value Decomposition of Jacobian $J(x)$ reveals that the singular values spectrum has a rapidly decreasing shape, with but a few large singular values. This shows that, indeed, the hidden representation learned to be sensitive to changes in only a few input-space directions, which is coherent with the hypothesis of the data concentrating along a low-dimensional manifold.

Note that while $J(x)$ contains all information to compute the sensitivity of $h = f(x)$ to movements in any input direction, performing a SVD yields a more directly informative orthonormal basis of directions, ordered from most to least sensitive. The subset of most sensitive directions in this basis can be interpreted as spanning the tangent space to the manifold at point $x$.

Rifai et al. (2011a) have successfully used these extracted "tangent" directions in a subsequent supervised classification training to encourage class prediction probabilities to be invariant to these directions, by using the tangent-propagation technique (Simard et al., 1992).

---

[2] What we will refer to as "the manifold", does not have to be a single connected sub-manifold of fixed dimensionality, but more generally a set of possibly disconnected stripes of varying dimensionality.

A Generative Process for Sampling Contractive Auto-Encoders

## 3. Generating samples along the manifold

Consider a pretrained CAE with hidden units $h_i = f_i(x)$, e.g., computed according to Eq. 1. Consider a local variation $\hat{x}$ of $x$ obtained by moving infinitesimally along the tangent plane defined by the CAE at $x$, i.e., with greater movement in the directions of the Jacobian $J$ associated with larger singular values:

$$\hat{x} = x + \sum_i^D \epsilon_i \frac{\partial f_i(x)}{\partial x}$$

where $\epsilon_i$ is a Gaussian perturbation in $h$-space with small variance $\sigma^2$, and the vector $\epsilon$ is isotropic with $\epsilon \sim N(0, \sigma^2 I_k)$.

We can now calculate the hidden units representation associated with the perturbed sample $\hat{x}$:

$$f_j(\hat{x}) = f_j\left(x + \sum_i^D \epsilon_i \frac{\partial f_i(x)}{\partial x}\right)$$

By using a first order Taylor expansion, taking advantage of the assumption that $\sigma$ is small and thus $\epsilon$ is small, we obtain a movement in the space of $h$ as follows:

$$f_j\left(x + \sum_i^D \epsilon_i \frac{\partial f_i(x)}{\partial x}\right) \approx f_j(x) + \sum_i^D \epsilon_i \frac{\partial f_i(x)}{\partial x}^T \frac{\partial f_j(x)}{\partial x}.$$

Hence, to first order, when we move $x$ along the directions captured by the CAE's Jacobian, it corresponds to a movement from $h = f(x)$ to $h + JJ^T\epsilon$, where $\epsilon$ is a small isotropic perturbation. This idea is exploited in Algorithm 1 below.

---

**Algorithm 1 : Sampling from a CAE.**
Let inputs $x \in [0,1]^d$ and representations $h \in [0,1]^k$ with encoding function $f : [0,1]^d \to [0,1]^k$ from input space to representation space, and decoding function $g : [0,1]^k \to [0,1]^d$ from representation space back to input space.

**Input:** $f$, $g$, step size $\sigma$ and chain length $T$
**Output:** Sequence $(x_1, h_1), (x_2, h_2), \ldots, (x_T, h_T)$
  Initialize $x_0$ arbitrarily and $h_0 = f(x_0)$.
  **for** $t = 1$ **to** $T$ **do**
    Let Jacobian $J_t = \frac{\partial f(x_t)}{\partial x_t}$.
    Let $\epsilon \sim N(0, \sigma I_k)$ isotropic Gaussian noise.
    Let perturbation $\Delta h = J_t J_t^T \epsilon$.
    Let $x_t = g(h_{t-1} + \Delta h)$ and $h_t = f(x_t)$
  **end for**

---

If the singular values of $J$ had a sharp drop-off, only movement on the manifold would be allowed. Furthermore, if the singular value spectrum was constant by parts (some large constant value for the first $m$ singular values, and 0 for the remaining ones), and in the limit where $\sigma \to 0$, then we claim that Algorithm 1 would do a random walk on the manifold. What if some singular values are non-zero or if $\sigma > 0$? We would take some steps "off" the manifold. However the reconstruction step (last step) of the algorithm would always bring us back towards the manifold (since $g$ is trained to output values on a set of high-density points, i.e., the training examples). We show below that in fact Algorithm 1 defines a stochastic process that asymptotically generates examples according to a well defined distribution. Note that the CAE used to compute $J$ does not have to be a single-layer CAE (as was the case in earlier CAE papers), and below we show experimental results with 2-layer CAEs that improve on single-layer CAEs. In that case, $J$ represents the Jacobian of the top-level layer with respect to the model's input.

**Theorem**: Algorithm 1 defines a stochastic process generating a sequence $h_1, h_2, \ldots$ that is an ergodic Harris chain with a stationary distribution $\pi$, so long as $J_t J_t^T$ is full rank.

**Proof**:

Note first that $\Delta h$ is a zero-mean Gaussian with full rank covariance matrix $\mathbb{E}[\Delta h \Delta h^T] = \mathbb{E}[J_t J_t^T \epsilon \epsilon^T J_t J_t^T] = J_t J_t^T \mathbb{E}[\epsilon \epsilon^T] J_t J_t^T = \sigma J_t J_t^T J_t J_t^T$. Since $\Delta h$ can be anywhere in $\mathbb{R}^k$, so can $h_{t-1} + \Delta h$. Let $\mathcal{H} = \{f(g(h)) : h \in \mathbb{R}^k\}$ the domain reachable in representation space when starting from anywhere in representation space and applying the decoder and encoder. This is the set in which all $h_t$ belong in Algorithm 1. Hence, by definition of $\mathcal{H}$, for any $h_t \in \mathcal{H}$, there exists $h \in \mathbb{R}^k$ such that $h_t = f(g(h))$, and since there exists $\delta > 0$ such that $\Delta h$ can take any value with probability density $p(\Delta h) > \delta$ so long as $\Delta h$ is in a bounded sphere around the origin, there exists $h_{t-1} \in \mathcal{H}$ such that $p(h_t|h_{t-1}) > \delta$.

The sequence of $h_t$'s defines a time-homogeneous Harris chain (a Markov chain with uncountable state) because it puts probability mass everywhere in $\mathcal{H}$, and it has a stochastic kernel $K$ s.t. $K(u, v) = p(h_{t+1} = v|h_t)$ and $K(u, V) = P(h_{t+1} \subset V|h_t = u)$ for a set $V$. In order for it to have a stationary distribution $\pi$ it is enough to show that it is irreducible, aperiodic and recurrent, all being true in virtue of the fact that, because $\mathcal{H}$ is a compact set, $\exists \delta > 0$ such that we can go from any element in $\mathcal{H}$ to any element in $\mathcal{H}$ in just one time step with probability greater than $\delta$: we can go from any region to any region in finite time (irreducibile chain), we can return to a just visited region (recurrent chain), and we can return to it in any number of time steps (aperiodic chain). These are the sufficient conditions for a stationary distribution $\pi$ to exist, i.e., such that $\pi(h') = \int K(h', h)\pi(h)dh$.$\square$

Since the sequence of $h_t$ converges in distribution, and $x_t = g(h_{t-1} + \Delta h)$ it can be similarly shown that the sequence of $x_t$'s also converges in distribution, i.e., that Algorithm 1 defines a Markov chain with a stationary distribution in the input space.

## 4. Efficiently training higher layers to be locally invariant to leading manifold directions

Previous work (Rifai et al., 2011a) has shown that the CAE's Jacobian characterizes a low dimensional tangent space that e.g., for images might correspond to small local deformations of the image such as translations. The CAE is able to learn these automatically instead of having to provide them as prior knowledge. Note that from the point of view of classification, the leading tangent space directions are often conceived as representing directions to which higher level features should be mostly invariant (assuming that each class is associated with a separate manifold, this is the "manifold hypothesis for classification" (Rifai et al., 2011a)).

Now if we follow the standard procedure for greedy layer-wise pre-training of deep networks, and stack a second similar CAE layer on top of the features learned by a first CAE layer, it is unlikely that this second CAE layer will learn to be highly invariant to these directions, to which the first layer is, on the contrary, very sensitive, as it would harm its reconstruction error. This behavior is to be contrasted with a very successful element of many deep architectures (especially for machine vision tasks), which is the addition of *pooling layers* in convolutional networks (LeCun et al., 1989). This powerful approach builds on prior knowledge of the 2D topology of images to pool the responses of several nearby receptive fields and yield units that are invariant, e.g., to small local translations of the input image. Since CAEs appear able to capture relevant local tangent directions without using prior domain knowledge, we wanted to find a way to automatically learn units that are, similarly, invariant to the corresponding deformations. This corresponds to *learning* a form of pooling (see also Coates and Ng (2011) for recent work on learning to pool). Hence we propose a slight addition to the training criterion of the second layer CAE to achieve this goal.

The normal CAE criterion for learning the second layer would be

$$\mathcal{J}_{\text{CAE}}(\theta') = \sum_{x \in \mathcal{D}} \left( L(f(x), g'(f'(f(x)))) + \lambda \|J'(f(x))\|^2 \right)$$

where $f'$ and $g'$ are defined similarly to $f$ and $g$ of the first layer, but using a different set of parameters $\theta' = \{W', b_{h'}, b_{r'}\}$, and where $J' = \frac{\partial f'(h)}{\partial h}$.

To encourage representation $h' = f'(h) = f'(f(x))$ to be most invariant to the directions captured by the first layer's $J$, we add a further term to the objective:

$$\mathcal{J}_{\text{CAE+}}(\theta') = \sum_{x \in \mathcal{D}} \left( L(f(x), g'(f'(f(x)))) + \lambda \|J'(f(x))\|^2 \right.$$
$$\left. + \lambda_p \mathbb{E}\left[ \|f'(f(x) + J(x)J(x)^T \epsilon) - f'(f(x))]\|^2 \right) \right.$$

where expectation $\mathbb{E}$ is over $\epsilon \sim \mathcal{N}(0, \sigma^2 I_k)$, and $\lambda_p \geq 0$ is a hyper-parameter. Note that $\Delta h = J(x)J(x)^T \epsilon$ is the same as was used in our CAE sampling procedure, which served as inspiration for this novel criterion. In practice, we will use a stochastic version of this criterion, where we sample a single $\epsilon$ each time we consider a different $x$.

Naturally this addition will likely worsen the achieved reconstruction error. But it is to be expected that as more abstract and invariant higher layers are learned, reconstruction from their representation alone will be further from the exact input.

The intended effect of this criterion is similar to Rifai et al. (2011a)'s motivation for using tangent-propagation with extracted "tangent" directions to yield class prediction probabilities that are invariant to these directions. There are however two major differences. First the criterion we propose is not tied to a supervised classification task, as it is local to the CAE+ layer. It can be viewed instead as a fully unsupervised learning of a form of pooling. Second, it does not require performing a SVD for each data point, and is thus much more efficient computationally.

## 5. Experiments

In this section, we will first evaluate empirically the sampling procedure defined in Section 3 for a trained CAE, by comparing it to samples generated by a trained Restricted Boltzmann Machines (RBM). Second we will show empirical results supporting the effectiveness of learning more invariant units with the criterion explained in section 4 in terms of classification performance. All our experiments were conducted on the MNIST, Caltech 101 Silhouettes, and the **Toronto Face Database (TFD)**. The latter is the largest dataset used for facial expression recognition. The dataset is composed of $100,000$ unlabeled

A Generative Process for Sampling Contractive Auto-Encoders

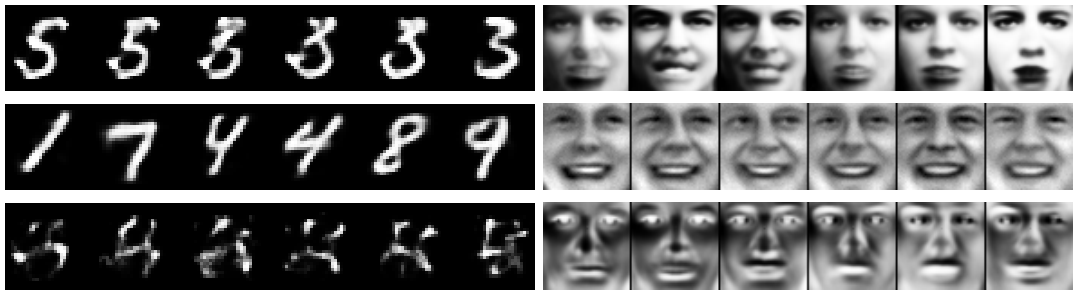

*Figure 1.* Samples from models trained on MNIST (left) and TFD (right). Top row: 2-layer CAE using proposed sampling procedure (Jacobian-based). Middle row: 2-layer DBN using Gibbs sampling. Bottom row: samples obtained by adding isotropic instead of Jacobian-based Gaussian noise.

images 48x48 pixels which makes it particularly interesting in the context of unsupervised learning algorithms, and 4000 labeled images with 7 facial expressions. We use the same preprocessing pipeline described in Ranzato et al. (2011).

### 5.1. Evaluating sample generation

We used the sampling procedure proposed in Section 3 to generate samples from two layer stacks of ordinary CAE (denoted CAE-2), that were trained on the MNIST and TFD datasets. To verify the importance of basing the stochastic perturbation of the hidden units on the CAE's Jacobian, we also run an alternative technique where we instead add isotropic noise. For comparison we also generated samples with Gibbs samling from a 2-layer Deep Belief Network denoted DBN-2 (i.e. stacking two RBMs). For the first RBM layer we used binary visible units for MNIST, and Gaussian visible units for TFD. Hidden units were binary in both cases. Figure 1 shows the generated samples for qualitative visual comparison. Figure 3 shows the evolution of the reconstruction error term, as we sample using either Jacobian-based or isotropic hidden unit perturbation. We see that the proposed procedure is able to produce very diverse samples of good quality from a trained CAE-2 and that properly taking into account the Jacobian is critical. Figure 2 shows typical first layer weights (filters) of the CAE-2 used to generate faces in Figure 1.

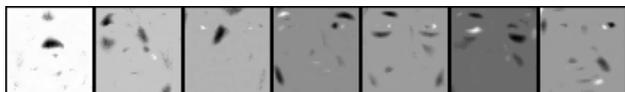

*Figure 2.* Typical filters (weight vectors) of the first layer from the CAE-2 used to produce face samples.

To get a more objective quantitative measure of the quality of the samples, we resort to a procedure proposed in Breuleux et al. (2011) that can be applied to compare arbitrary sample generators. It consists in measuring the log-likelihood of a test set (not used to train the samplers) under the density obtained from a Parzen-Windows density estimator[3] based on 10000 generated samples. Table 1 shows the thus measured log-likelihoods.

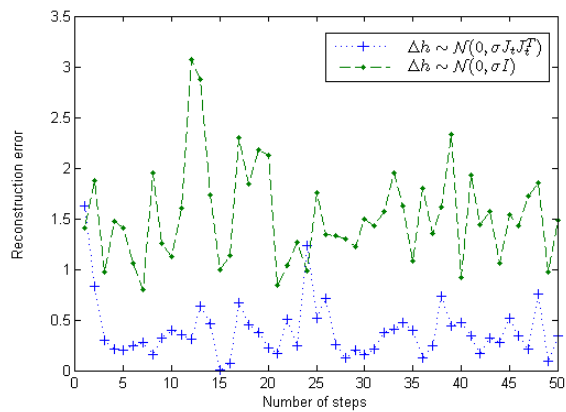

*Figure 3.* Evolution of the reconstruction error term, as we sample from CAE-2 trained on MNIST, starting from uniform random pixels (point not shown, way above the graph). Sampling chain using either Jacobian-based (blue) or isotropic (green) hidden unit perturbation. Reconstruction error may be interpreted as an indirect measure of likelihood.

*Table 1.* Log-Likelihoods from Parzen density estimator using 10000 samples from each model

|  | DBN-2 | CAE-2 |
|---|---|---|
| TFD | **1908.80** $\pm$ 65.94 | **2110.09** $\pm$ 49.15 |
| MNIST | **137.89** $\pm$ 2.11 | 121.17 $\pm$ 1.59 |

### 5.2. Evaluating the invariant-feature learning criterion

Our next series of experiments investigates the effect of the training criterion proposed in section 4 to learn

---

[3]using Gaussian kernels whose width is cross-validated on a validation set

A Generative Process for Sampling Contractive Auto-Encoders

more invariant second-layer features. The corresponding model is denoted CAE-$2_p$.

**How to measure invariance to transformations known a-priori?** In order to validate that the proposed criterion does learn features that are more invariant to transformations of interest, we define the following *average normalized sensitivity* score, following the ideas put forward in Goodfellow et al. (2009). Let $T(x)$ represent a random deformation of $x$ in directions of variability known a priori, such as affine transformations of the ink in an image (corresponding for example to small translations, rotations or scaling of the content in the image). Then $(f_i(x) - f_i(T(x)))^2$ measures how sensitive is unit $i$ to $T$ (when large) or how invariant to it (when small) it is. Following Goodfellow et al. (2009) we also need some form of normalization to account for features that do not vary much (maybe even constant): we will normalize the sensitivity of each unit by $V[f_i]$, the empirical variance of $f_i(x)$ over the data set. This yields for each unit $i$ and each input $x$ a normalized sensitivity

$$s_i(x) = \frac{(f_i(x) - f_i(T(x)))^2}{V[f_i]}.$$

Like Goodfellow et al. (2009), we focus on a fraction of the units, here the 20% with highest variance, and define the normalized sensitivity $\gamma(x)$ of the *layer* for an example $x$ as the average of the normalized sensitivity of these selected units. Finally, the *average normalized sensitivity* score $\bar{\gamma}$ is obtained by computing the average of $\gamma(x)$ over the training set. When comparing layers learned by two different models $A$ and $B$, statistical significance of the difference between $\bar{\gamma}_A$ and $\bar{\gamma}_B$ is assessed by computing the standard error[4] of the mean of differences $\gamma_A(x) - \gamma_B(x)$.

**Experimental comparison of sensitivity to a-priori known deformations.** We considered stochastic affine deformation $T(x)$ for MNIST test-set digits, controlled by a set of 6 random parameters. These produced slightly shifted, scaled, rotated or slanted variations of the digits.

Table 2 compares the resulting *average normalized sensitivity* score obtained by the second layer learned by the CAE-$2_p$ algorithm, to that obtained for several alternative models. These results confirm that the CAE-$2_p$ has learned features that are significantly more invariant to this kind of deformations, even though they have not been explicitly used during training. DBN-2 is a Deep Belief Network obtained by stacking two RBMs.

Table 2. Average normalized sensitivity ($\bar{\gamma}$) of last layer to affine deformations of MNIST digits. The deformations were not used in any way during training. Second column shows difference with CAE-$2_p$ together with standard error of the differences. The proposed CAE-$2_p$ appears to be *significantly* less sensitive (more invariant) than other models.

| Model | $\bar{\gamma}$ | $\bar{\gamma} - \bar{\gamma}_{\text{CAE}-2_p}$ |
|---|---|---|
| CAE-1 | 8.23 | 6.28 ± 0.04 |
| CAE-2 | 2.84 | 0.89 ±0.08 |
| DBN-2 | 2.36 | 0.41 ±0.025 |
| CAE-$2_p$ | **1.95** | 0 |

**Effect of invariant-feature learning on classification performance.** Next we wanted to check whether the ability to learn more invariant features of the CAE-$2_p$ could translate to better classification performance. Table 3 shows classification performance on the TFD dataset of a deep neural network pretrained using several variants of CAEs, and then fine-tuned on the supervised classification task. For comparison we also give the best result we obtained with a non-pretrained multi-layer perceptron (MLP). We see that the criterion for learning more invariant features used in CAE-$2_p$ yields a significant improvement in classification. We get the best performance among methods that do not explicitly use prior domain knowledge.[5]

Table 3. Test classification error of several models, trained on TFD, averaged over 5 folds (reported with standard deviation).

| MLP | CAE-1 | CAE-2 | CAE-$2_p$ |
|---|---|---|---|
| 26.17 ± 3.06 | 24.12 ± 1.87 | 23.73 ± 1.62 | **21.78** ± 1.04 |

## 6. Future Work and Conclusion

We have proposed a converging stochastic process which exploits what has been learned by a CAE to generate samples, and we have found that these samples are not only visually good, they "generalize" well, in the sense of populating the space in the same areas where one tends to find test examples. A related idea has also allowed us to train the second layer of a 2-layer CAE, acting like pooling or invariance-seeking features, and this has yielded improvements in classi-fication and invariance.

---

[4]This is a slight approximation that considers the $V[f_i]$ as constants.

[5]for comparison, Ranzato et al. (2011) reports that SVM achieve 28.5% and sparse coding : 23.4%. Better performance is obtained by convolutional architectures (17.6%) but the convolutional architecture and hard-coded pooling uses prior knowledge of the topology of images.

The invariance criterion for training pooling units proposed here works well for feature extraction for classification problems (and so do classical pooling strategies) but is not optimal for pure unsupervised learning. Indeed, it throws away some of the information along the first layer's leading direction of variation (which are typically not useful for classification, but would be useful to reconstruct the input). A more general approach that we propose to investigate is based on the idea of learning a representation that does not throw these variations away but instead *disentangles* them from each other (Bengio, 2009). We believe this could be achieved by criteria which encourage the active features to be invariant to the input variations captured by *other* features.

Another direction of future work regards linking the CAE to a density model giving rise to the appropriate local covariance (i.e., the one used in the generative process). If we consider a small neighborhood around a point $x$ with density $p(x)$, then the local variance is the variance associated with the density that is the product of $p$ with a local kernel (such as as Gaussian or uniform ball) centered at $x$. The generating process we have proposed here essentially corresponds to moves according to a Brownian motion in which the local mean and covariance are location-dependent. The interesting question is how one should pick those local means and covariances so as to replicate a target density. Intuitively, these should depend on the local structure of the density, so that the mean moves points towards the manifold (in the direction of the log-likelihood gradient) and the covariance disallows moving out of the manifold. Note that both of these are captured by the CAE.